\documentclass[10pt,twocolumn,letterpaper]{article}

\usepackage{cvpr}
\usepackage{times}
\usepackage{epsfig}
\usepackage{graphicx}
\usepackage{amsmath}
\usepackage{amssymb}
\usepackage{algorithm}
\usepackage{algorithmic}

\usepackage[pagebackref=true,breaklinks=true,letterpaper=true,colorlinks,bookmarks=false]{hyperref}

\cvprfinalcopy 

\def\cvprPaperID{729} 

\ifcvprfinal\fi
\begin{document}

\title{Unsupervised Learning for Large-Scale Fiber Detection and Tracking in Microscopic Material Images}

\author{Hongkai Yu\textsuperscript{$1$}, Dazhou Guo\textsuperscript{$1$}, Zhipeng Yan\textsuperscript{$2$}, Wei Liu\textsuperscript{$3$}, \\Jeff Simmons\textsuperscript{$4$}, Craig P. Przybyla\textsuperscript{$4$}, Song Wang\textsuperscript{$1$}\thanks{E-mail: songwang@cec.sc.edu} 
\\
\textsuperscript{$1$}University of South Carolina, Columbia, SC, 
\textsuperscript{$2$}University of California, San Diego, CA \\
\textsuperscript{$3$}Tencent AI Lab, Shenzhen, China, 
\textsuperscript{$4$}Air Force Research Lab, Dayton, OH \\}


\maketitle

\begin{abstract}

Constructing 3D structures from serial section data is a long standing problem in microscopy. The structure of a fiber reinforced composite material can be  reconstructed using a tracking-by-detection model. Tracking-by-detection algorithms rely heavily on detection accuracy, especially the recall performance. The state-of-the-art fiber detection algorithms perform well under ideal conditions, but are not accurate where there are local degradations of image quality, due to contaminants on the material surface and/or defocus blur. Convolutional Neural Networks (CNN) could be used for this problem, but would require a large number of manual annotated fibers, which are not available. We propose an \textbf{unsupervised learning} method to accurately detect fibers on the large scale, that is robust against local degradations of image quality. The proposed method \textbf{does not require manual annotations}, but uses fiber shape/size priors and spatio-temporal consistency in tracking to simulate the supervision in the training of the CNN. Experiments show significant improvements over state-of-the-art fiber detection algorithms together with advanced tracking performance.       
\end{abstract}

\section{Introduction}

Continuous fiber reinforced composite materials are desired in aerospace applications because of their superior strength and stiffness, as compared with traditional materials~\cite{Knox2001, Suresh2013, zhou2016large}. Microstructure characterization is performed in fiber reinforced composites, as well as most other materials, as a means of controlling their properties. In particular, the 3-dimensional (3D) structures are important to characterize because not all topological properties can be inferred from 2-dimensional sections alone. Reconstruction of the 3D structures is desirable and helpful in material science.

\begin{figure}[htbp]
\begin{centering}
\includegraphics[width=1.0\columnwidth]{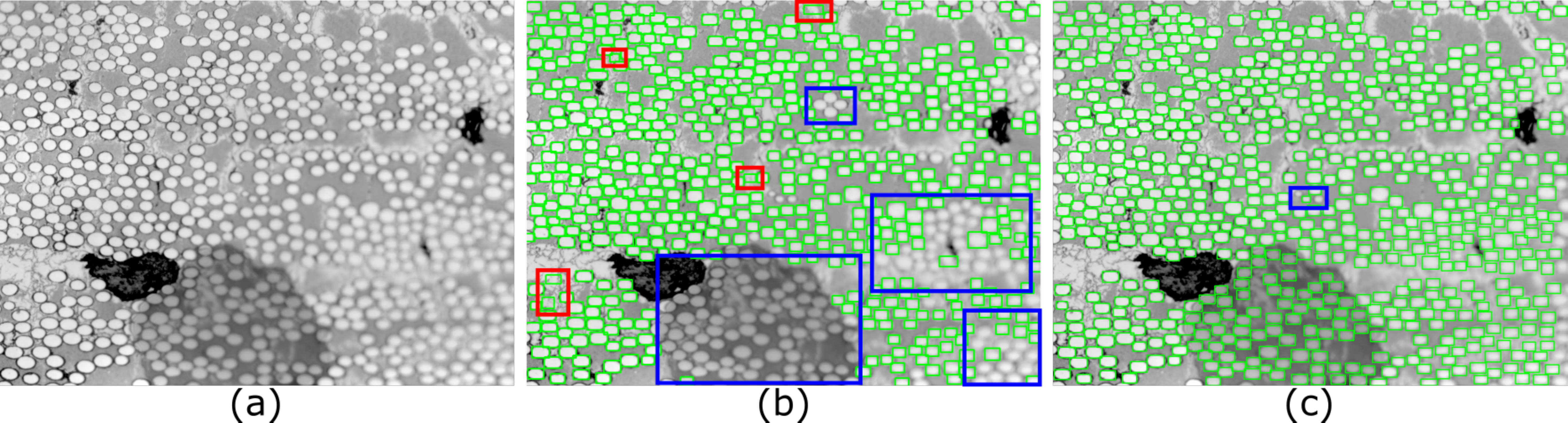}
\par\end{centering}
\caption{Comparison of the state-of-the-art method of~\cite{zhou2016large} and the proposed method: (a) a sectioned microscopic material image, (b) fiber detections by~\cite{zhou2016large}, and (c) fiber detections by the proposed method. Detected fibers are marked by green bounding boxes. Red and blue boxes highlight the false positive and false negative errors respectively. Note that \cite{zhou2016large} fails where the image quality is poor, \eg, blurred and stained regions.}
\label{fig:motivation}
\end{figure}

While 3D microstructures may be characterized with Transmission Electron Tomography~\cite{venkatakrishnan2013model,batenburg20093d} or X-ray Tomography~\cite{williams2010damage}, the size scale of the SiC fibers used in this study renders them opaque to electron and X-ray beams, so 2D optical sections were made of bulk samples. Recent studies~\cite{yu2016groupwise, zhou2016large} showed that the 3D fiber structures could be reconstructed by tracking the detected fibers through the 2D image sequence, which is modeled as a tracking-by-detection problem in computer vision. The  tracking performance of tracking-by-detection algorithms is largely dependent on the detection accuracy, especially the recall performance~\cite{hong2016online}. More accurate object detections could greatly improve tracking-by-detection algorithms.

State-of-the-art fiber detection is currently achieved by~\cite{zhou2016large}, which uses the physics-based knowledge that the shape of the sectioned fiber is approximately elliptical. In~\cite{zhou2016large}, the authors first apply the EMMPM segmentation algorithm~\cite{comer2000emmpm}, which is a Markov Random Fields based unsupervised algorithm for image segmentation, to segment the material images into fiber and non-fiber regions and then utilize a Hough transform based ellipse fitting algorithm~\cite{xie2002new} to detect the fibers. The minimum bounding boxes of each detected ellipse are taken as the fiber detection result. This algorithm performs well if material image quality is good in the neighborhood of the fiber, but it performs poorly when the image quality is bad. 

In the present case, the images were degraded in local regions because of contaminants. During sample preparation, the samples are ground, cleaned, and then imaged. The cleaning was performed with a water rinse that was subsequently dried. Sometimes, during the drying process, liquid water in the form of drops was left on the surface. This creates darker areas with thickness fringes. Local areas of defocus blur are also quite common when the surface being imaged is not flat, leaving areas out of focus. These cases result in degraded areas in the images. In the degraded situations, the state-of-the-art fiber detection algorithm~\cite{zhou2016large} fails to accurately segment the fiber regions, resulting in false positive or false negative errors. This is shown in Fig.~\ref{fig:motivation}. 

When a large number of manual annotations are available, CNN can accurately detect various objects~\cite{girshick2015fast,ren2015faster,qin2016joint}, even for poor quality or blurred images~\cite{zhang2016faster,li2017scale}. Supervised learning algorithms using CNN are robust and accurate but require manual annotations. In the present study, each single image contains approximately 600 fibers. Manually annotating so many fibers for multiple images is tedious and time consuming. 

In this paper, we propose an unsupervised learning approach that accurately detects large-scale fibers without the need for manual annotations. In one image sequence, we find poor imaging conditions occur occasionally and locally, but the majority of the data is clear and trackable. Since accurate detection is required in regions of poor imaging conditions, it is necessary to develop accurate estimates of the locations of the fibers in these regions. For this purpose, the spatio-temporal consistency in fiber tracking is applied to estimate these false-negative detections. In addition, the spatio-temporal consistency in fiber tracking could also remove some false-positive detections. 

The basic idea here is to use fiber shape/size priors together with the spatio-temporal consistency in tracking to simulate the supervision during the training of a CNN. In this way, an unsupervised CNN approach was developed for which experimental results show a significant improvement over the performance of the state-of-the-art fiber detection of~\cite{zhou2016large}, coupled with advanced tracking. 

The remainder of the paper is organized as follows. Section~\ref{sec:related_work} briefly reviews the related work. Section~\ref{sec:our_method} describes the proposed fiber detection and tracking method. Quantitative and qualitative evaluation results and discussions are presented in Section~\ref{sec:experiment}, followed by brief conclusions in Section~\ref{sec:conclusions}.
\section{Related work}\label{sec:related_work}
\textbf{Fiber detection and tracking:} With the prior knowledge of fiber shape, some unsupervised methods~\cite{yu2016groupwise,zhou2016large,puatruaucean2012parameterless}, using ellipse detection, are proposed to detect the large-scale fibers in material images. However, these previous unsupervised methods using low-level features are not robust in the degraded  material images. Recently, advanced CNN based object detectors~\cite{girshick2015fast,ren2015faster,Hu_2017_CVPR} could also be applied for large-scale fiber detection, while the requirement of manual annotations for training CNN is expensive. Fiber tracking can be easily addressed if the inter-slice distance is small during the material cross-sectioning. Many tracking-by-detection methods can be utilized to track the large-scale fibers~\cite{yu2016groupwise,zhou2016large,Milan2014}.    

\textbf{Unsupervised learning with CNN:} In this paper, unsupervised learning refers to learning without any manual annotations. With several prior knowledge or constraints, the powerful CNN can be used for unsupervised learning,  approaching approximately close performance with the CNN trained with full supervision~\cite{li2016unsupervised,goroshin2015unsupervised,
Zhang_2017_ICCV,Lee_2017_ICCV,Croitoru_2017_ICCV,wang2015unsupervised}. Due to the lack of manual annotations, different video or multiple images based properties are frequently utilized to simulate the human supervisions. Optical flow based motion information is used to assist CNN for edge detection
~\cite{li2016unsupervised}. Assuming that adjacent video frames contain similar representation, feature learning is performed in unlabeled video data~\cite{goroshin2015unsupervised}. The fusion of multiple saliency maps is used to simulate human supervision to train CNN without manual annotations to improve unsupervised saliency detection~\cite{Zhang_2017_ICCV}. Using the chronological order of frames as supervision, unsupervised deep representation Learning is applied~\cite{Lee_2017_ICCV}. Given unlabeled videos, unsupervised object discovery is used to train a CNN for detecting objects in single images
~\cite{Croitoru_2017_ICCV}. 

\textbf{Spatio-temporal consistency:} Spatio-temporal consistency, as an important property in video processing, has many vision applications such as video object proposals~\cite{oneata2014spatio}, object instance search in videos~\cite{meng2016object}, human segmentation~\cite{liang2017learning}, etc. Assuming tracked patches have similar visual representation in deep feature space, unsupervised learning of visual representations is accomplished~\cite{wang2015unsupervised}. As described in~\cite{Feichtenhofer_2017_ICCV}, tracking and detection can be jointly carried out in a supervised CNN based framework with manual annotations.  
  
Inspired by these researches, we expect that fiber tracking and fiber detection could 
work collaboratively to achieve better performance in an unsupervised manner. Without any manual annotations, we combine fiber shape/size priors and spatio-temporal consistency in tracking to simulate the human-like supervision in training a CNN based object detector, providing effective fiber detection and tracking simultaneously. 
\section{Proposed method}\label{sec:our_method}
\begin{figure}[htbp]
\begin{centering}
\includegraphics[width=1.0\columnwidth]{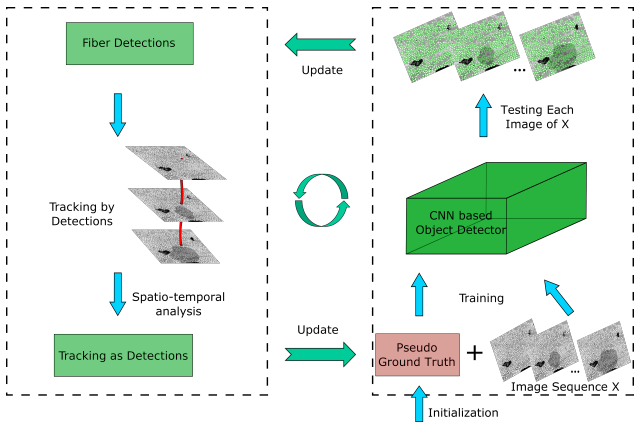}
\par\end{centering}
\caption{The framework of the proposed unsupervised learning method for large-scale fiber detection and tracking. Left part: fiber tracking, Right part: CNN training/testing. The input is an image sequence without any manual annotations.}
\label{fig:framework}
\end{figure}

In this paper, we propose an unsupervised learning method to accurately detect  and track the large-scale fibers. The input is an image sequence without any manual annotations. With some unsupervised methods~\cite{zhou2016large,zitnick2014edge} using shape or size prior, the initial pseudo ground truth of fiber detections could be obtained. The powerful CNN based object detector is used as the base detector in our framework. The spatio-temporal consistency in fiber tracking is analyzed to simulate the supervision to correct and refine the pseudo ground truth (reduce false-positive and false-negative detection errors). Refined pseudo ground truth would train a better CNN based object detector, and the improved object detector would generate more accurate fiber detections so as to boost fiber tracking, while better fiber tracking would further correct and refine the pseudo ground truth. We expect that the CNN based object detector and tracking-by-detection algorithm could help each other. For a robust solution, the processes of CNN training/testing and fiber tracking are performed alternately in several iterations. The diagram of the framework of the proposed method is illustrated in Fig.~\ref{fig:framework}. For the CNN based object detector, we use Faster R-CNN~\cite{ren2015faster} (Region-based Convolutional Neural Networks) due to its outstanding detection performance. For the fiber tracking, we use Kalman filter based fiber tracking algorithm~\cite{zhou2016large} because of its satisfactory performance in fiber tracking. We will introduce the details of each part in the following. 
 
\subsection{Initialization}
The proposed method needs an initial pseudo ground truth for initialization. We give two unsupervised methods for accomplishing this: 1) the EdgeBox~\cite{zitnick2014edge} algorithm and 2) the algorithm by~\cite{zhou2016large}, which we refer to in this paper as the state-of-the-art fiber detection algorithm or by the acronym EMMPMH\footnote{The acronym stands for EMMPM segmentation with Hough detection.}.

The EdgeBox algorithm detects contour-representative object proposals, including both fiber and non-fiber regions. To reduce the detection errors, we use a size prior to eliminate false positives. This is accomplished by, first computing the mean 
size/area, $a$, of the fibers in a sample image of the input image sequence that were detected by EMMPMH~\cite{zhou2016large}. We then prune the object proposals whose sizes are out the range of $[0.2a, 2a]$ in the input image sequence, followed by a Non-Maximum Suppression (NMS). The NMS threshold is set to 0.1 because of the highly overlapped proposal regions given by EdgeBox. The state-of-the-art fiber detection algorithm EMMPMH~\cite{zhou2016large} detects ellipse-like objects. It takes only one input: the number of classes for the segmentation algorithm, which we set to $3$. 

\begin{figure}[htbp]
\begin{centering}
\includegraphics[width=1.0\columnwidth]{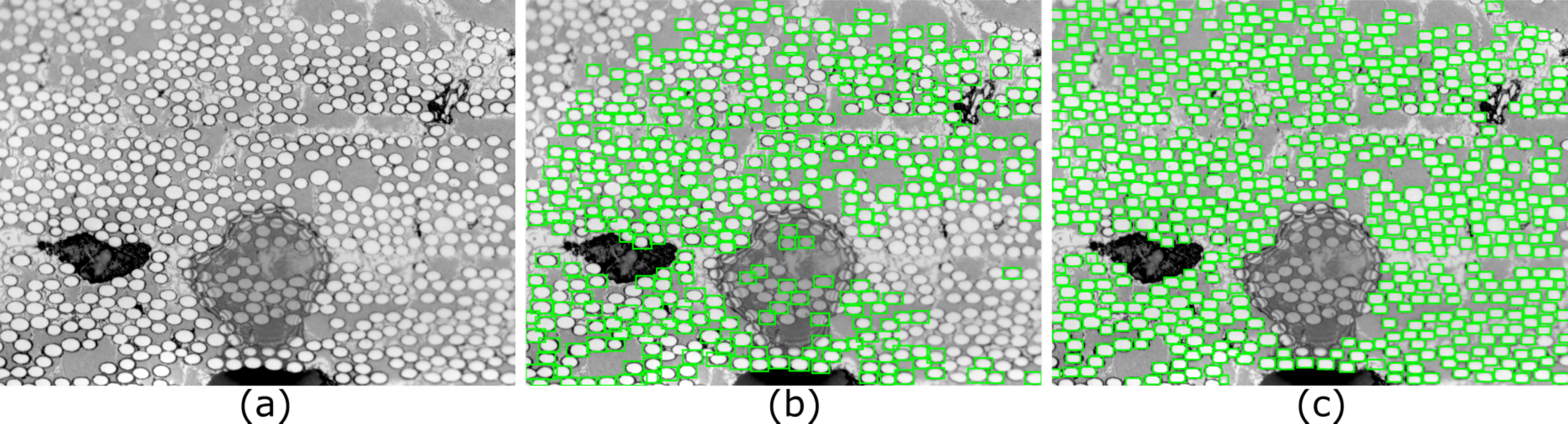}
\par\end{centering}
\caption{Initialization for the pseudo ground truth of fiber detections shown as green bounding boxes. (a) original image, (b) initialization by EdgeBox~\cite{zitnick2014edge} showing many missed detections, and (c) initialization by EMMPMH~\cite{zhou2016large} with fewer missed detections.}
\label{fig:Initialization}
\end{figure}

Both EdgeBox and EMMPMH algorithms are unsupervised image processing methods using either size or shape prior to detect fibers. Both suffer from false positive and false negative identifications, as shown in Fig.~\ref{fig:Initialization}. Because the EMMPMH algorithm utilizes more specific shape prior, it generates better initialization results than the EdgeBox algorithm. We used both methods in the experiments (see below). The proposed method outperformed both of these algorithms.

\subsection{Faster R-CNN for fiber detection}

Recently, the Faster R-CNN is developed and has superior performance for many object detection related tasks~\cite{qin2016joint,li2017scale}. Because it is stable and efficient, it is utilized as the CNN based object detector in the proposed method. Given the pseudo ground truth, the Faster R-CNN can be trained for fiber detection.
Faster R-CNN is composed of two modules: 1) a deep convolutional network that proposes regions (the Region Proposal Network-RPN) and 2) a Fast R-CNN detector ~\cite{girshick2015fast} that uses the proposal regions for object detection and classification. Since the RPN shares full-image convolutional features with the detection network, the computation cost of region proposals is low. Essentially, the RPN serves as the `attention' mechanism, telling the Fast R-CNN detector where to look.


In order to handle different scales and aspect ratios of objects, the Faster R-CNN introduces anchors of different scales and aspect ratios in a sliding window manner.  In the proposed method, we use 5 scales ($32^2, 64^2, 128^2, 256^2, 512^2$ pixels) and 3 aspect ratios ($1:1$, $1:2$, and $2:1$). Following~\cite{ren2015faster}, the anchors whose Intersection-over-Union (IoU) overlaps with a pseudo ground-truth box are above 0.7 or below 0.3 are set as positive and negative samples respectively during training RPN. The loss function $L$ of Faster R-CNN contains two components: 
\begin{equation}
L= L_{cls} + \lambda L_{reg}, 
\label{eq:loss}
\end{equation}
where $L_{cls}$ is the normalized \textit{classification} loss and $L_{reg}$ is the normalized \textit{regression} loss with a balance weight $\lambda$. Here, $\lambda$ was set to $1$, following~\cite{girshick2015fast}. $L_{cls}$ is a log loss over two classes (fiber v.s. non-fiber) and $L_{reg}$ is the smooth L1 loss over bounding box locations~\cite{girshick2015fast}. Same as~\cite{ren2015faster}, we sample 256 anchors (positive and negative) for one image during training RPN (first module). For training Fast R-CNN (second module), we fix the IoU threshold for NMS as 0.7 so as to generate about 2,000 proposal regions per image. The VGG network~\cite{simonyan2014very} is used as the base convolutional layers to extract deep features. The whole Faster R-CNN is a unified network that can be trained end-to-end by back propagation and stochastic gradient descent. 

    
\subsection{Tracking by detections} \label{sec:tracking-by-detection}
Given the large-scale detected fibers by Faster R-CNN, we model this problem as a tracking-by-detection problem in the image sequence. Since Kalman filter has been proven as a reliable model for large-scale fiber tracking in~\cite{yu2016groupwise,zhou2016large}, we apply Kalman filter to track each fiber by recursively performing prediction, association and correction along the image sequence.

For later fiber detection, we define the tracking state $\textbf{s} = (x_1, y_1, v_{x_1}, v_{y_1}, x_2, y_2, v_{x_2}, v_{y_2})^T$ to denote the tracked fiber in the 2D images, where the first half is for the top-left point of the fiber's bounding box and the last half is for the bottom-right point of the fiber's bounding box. $(x,y)^T$ is the location and $(v_{x},v_{y})^T$ is the velocity in horizontal and vertical directions. We set up a Kalman filter to track each fiber and assume that each fiber is smooth in 3D space with a constant velocity. This means that the tracking state evolves linearly from image to image. The prediction and correction steps are the same as those in the traditional Kalman filter. During the association step, we use the Hungarian algorithm~\cite{Kuhn1955} for a minimum-total-distance bipartite matching between the centers of the bounding boxes of the predicted fibers and those of the detected fibers. The numbers of predictions and detections are usually different, so dummy nodes are introduced into the Hungarian algorithm and the distance to a dummy node is set to 100 pixels in our experiments.

%

\subsection{Tracking as detections} \label{sec:spatio-temporal}

\begin{figure*}[htbp]
\begin{centering}
\includegraphics[width=1.0\textwidth]{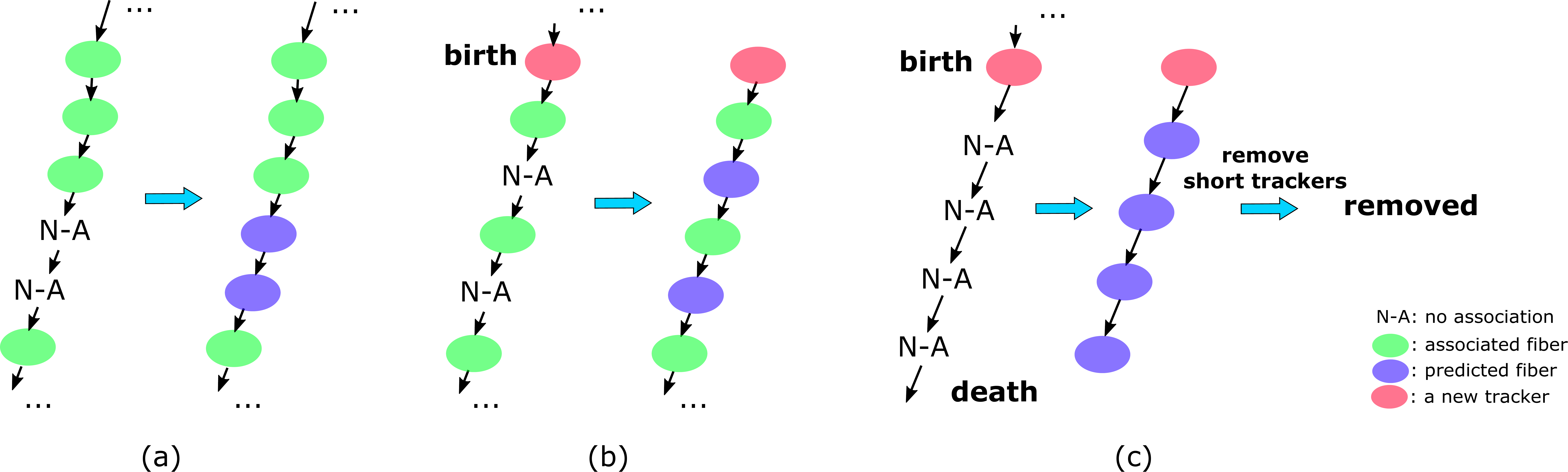}
\par\end{centering}
\caption{The unsupervised strategy of spatio-temporal analysis in fiber tracking. (a) making up for false-negative detections by predictions (added), (b) tracking birth by a true positive detection (saved), and (c) tracking birth and death by a false-positive detection (removed).}
\label{fig:spatio-temporal}
\end{figure*}

After fiber tracking, if we ignore the tracking identities and simply take the tracked bounding boxes as the detected fibers. Thus, an updated set of fiber detections is  obtained on each image of the image sequence. These detections are not perfect, since errors might occur during tracking that introduce false positives or false negatives. An unsupervised spatio-temporal analysis strategy is developed to reduce these errors. This section describes this strategy. 

\textbf{Spatio-temporal analysis:} Two complications are encountered here: 1) detected fibers in an image for which there are no trackers in previous images and 2) tracking drift when trackers are not associated with any fiber in several subsequent images. In order to correct these problems, tracking births and deaths are included in the algorithm. For the birth algorithm, we start a new Kalman filter to create a tracker  for each detected fiber that is not associated with any predictions of current set of Kalman filters. For the death algorithm, the Kalman filter is stopped if it has moved out of the image boundary or it has not been associated for a continuous sequence of $\alpha$ images. Note that the un-associated fibers introduce their predictions as tracked locations for continuous $\alpha$ images before the Kalman filter is stopped. This might lead to some erroneous detections being introduced. For this reason, after tracking, we prune the tracked fibers whose trajectories are shorter or equal than $\beta$, followed by a NMS on each image. We set $\alpha=5$ and $\beta=5$ in our experiments. NMS threshold for EMMPMH initialized tracking is set to 0.7 and for EdgeBox initialized tracking is set to 0.1. 

The unsupervised strategy of spatio-temporal analysis is shown in Fig.~\ref{fig:spatio-temporal}. For the unsupervised strategy of spatio-temporal analysis, we assume that 1) some of the missed detections can be added back by tracking predictions, 2) it is highly possible that true positive detections will be tracked through more than $\beta$ images, 3) false positive detections might not be associated through a continuous sequence of $\alpha$ images. This unsupervised strategy is reasonable for fiber tracking in the cross-sectioned 3D material sample, however it is still not perfect suffering from different  detection and tracking errors, therefore we tend to run the Faster R-CNN and fiber tracking algorithms alternately for several iterations in order to obtain improved fiber detection and tracking simultaneously. The detailed steps of the proposed algorithm are summarized in Algorithm~\ref{alg1}. In our experiment, the proposed algorithm always converged in 3 to 4 iterations. After convergence, a well trained Faster R-CNN model is built and can be applied to directly detect fibers on a new material image without performing fiber tracking.

\begin{algorithm}\caption{Unsupervised Learning for Large-Scale Fiber Detection and Tracking}
	\textbf{Input:} a sequence of microscopic material images  without any manual annotations, denoted as \textbf{X}. 
	\newline\textbf{Output:} fiber detection and tracking on each image of \textbf{X}, and a well trained Faster R-CNN model.
	\vspace{-1em}
	\begin{algorithmic}[1]
		\STATE Initialize the pseudo ground truth $\mathbf{G_p}$ of fiber detections using~\cite{zhou2016large} or~\cite{zitnick2014edge}  on each image of \textbf{X}.
		\REPEAT
		\STATE  Train a Faster R-CNN from scratch using $\mathbf{G_p}$ if previous Faster R-CNN model is not available. Otherwise, fine-tune the previous Faster R-CNN model using $\mathbf{G_p}$.  		 
		\STATE  Apply the trained Faster R-CNN on each image of \textbf{X} to detect fibers as $\textbf{D}$ and save it. 
		\STATE  Track detected fibers $\textbf{D}$ on \textbf{X} as in 
		Section~\ref{sec:tracking-by-detection} and save it. 
		\STATE  Take the tracked fibers as detections with the spatio-temporal analysis as in Section~\ref{sec:spatio-temporal}. 
		\STATE	Update the pseudo ground truth $\mathbf{G_p}$.
		\STATE	Save the current Faster R-CNN model.
		\UNTIL{convergence or maximum iterations reached}
	\end{algorithmic}
	\label{alg1}
\end{algorithm} 
\section{Experimental results}\label{sec:experiment}
In the experiment, we apply the proposed method to detect and track large-scale fibers from S200, an amorphous SiNC matrix reinforced by continuous Nicalon fibers. The microscopic images were collected by the RoboMet.3D automated serial sectioning instrument~\cite{robomet}. It takes about 15 minutes to grind for one slice. Given a material sample of S200, RoboMet.3D cross-sections the sample by mechanical polishing with dense inter-slice distance 1 $\mu m$, and each slice was then imaged with an optical microscope. We collect three datasets, denoted as `Set1', `Set2' and `Set3', to evaluate the proposed method. Set1 is a sequence of 90 images and $40\%$ of images contain certain-level degraded situations such as blurred and stained regions,  as shown in Fig.~\ref{fig:motivation}(a). Set2 is a sequence of 50 images and $30\%$ of images have certain-level degraded situations. Set3 is a set of 99 single images and $15\%$ of images have certain-level degraded situations. The size of each image is $1292\times968$ and each image contains about 600 fibers. 

The proposed method described in Algorithm~\ref{alg1} does not need any manual annotations, but we manually annotate the corresponding ground truth only for the evaluation purpose. We run Algorithm~\ref{alg1} (with tracking) on the image sequence Set1 without manual annotations, and obtain a well trained Faster R-CNN model as 
$\mathbf{M^{1}_{EMMPMH}}$ using EMMPMH~\cite{zhou2016large} initialization and another well trained Faster R-CNN model as $\mathbf{M^{1}_{EdgeBox}}$ using EdgeBox~\cite{zitnick2014edge} initialization. Running  Algorithm~\ref{alg1} (with tracking) on Set2 without manual annotations, we could obtain a well trained Faster R-CNN model as $\mathbf{M^{2}_{EMMPMH}}$ using EMMPMH initialization and another well trained Faster R-CNN model as $\mathbf{M^{2}_{EdgeBox}}$ using EdgeBox initialization. The well trained Faster R-CNN models on one dataset are then respectively applied to detect the large-scale fibers on each single image on another two datasets without tracking. On the collected Set1, we manually annotate the bounding boxes of fibers on each image as the ground truth for detection evaluation and link them across the image sequence as the ground truth for tracking evaluation. On the collected Set2 and Set3, we manually annotate the bounding boxes of fibers on each image for detection evaluation only. For the detection ground truth, we label all the fibers on each image. For the tracking ground truth, we label as many as we can, leading to 481 fibers' trajectories along the Set1.   

In our experiment, the maximum iteration in Algorithm~\ref{alg1} is set to 4. Within each iteration, we train Faster R-CNN for 10 epochs. The learning rate is 0.001 and the batch size is 2 images during training. For Algorithm~\ref{alg1}, we try two kinds of initializations for pseudo ground truth: EMMPMH and EdgeBox. We denote the proposed Algorithm~\ref{alg1} using the initialization EMMPMH as `Proposed-EMMPMH' and the initialization EdgeBox as `Proposed-EdgeBox'. After obtaining the well trained Faster R-CNN model $\mathbf{M}$, we denote directly applying the well trained model on single images (without tracking) to detect large-scale fibers as `Proposed-$\mathbf{M}$'. We use MXNet to implement the code of Faster R-CNN framework. With a GeForce GTX 1080Ti GPU and a 12-core CPU, it takes about half an hour to run one iteration (Faster R-CNN training plus large-scale fiber tracking) in Algorithm~\ref{alg1} with Set1 (a 90-slice image sequence) as input and only takes about 0.2 seconds to detect the large-scale fibers on one material image when testing the trained Faster R-CNN model.      

Five metrics are used to evaluate the fiber detection performance on Set1, Set2 and  Set3: Precision, Recall, F-measure, Number of False Positives per image ($N_{fp}$ per image), and Number of False Negatives per image ($N_{fn}$ per image). For all the methods, we use a uniform threshold of 0.5 for the IoU between the predicted bounding box and ground truth. Because each image contains large-scale fibers (about 600), percentage results might be not representative enough to display errors. Therefore, we also show $N_{fp}$  per image and $N_{fn}$ per image to illustrate the detection errors. Higher (Precision, Recall and F-measure) and lower ($N_{fp}$ and $N_{fn}$ per image) indicate the better detection performance. In our experiment, an ellipse detection algorithm ELSD~\cite{puatruaucean2012parameterless} is used as another comparison method for fiber detection, together with the above mentioned EMMPMH and EdgeBox methods. All these three comparison methods are unsupervised and do not need manual annotations. For ellipse detections by EMMPMH and ELSD, we take the minimum bounding boxes of each detected ellipse as their outputs. In addition, we also evaluate the fiber tracking performance on Set1 in terms of five widely used metrics~\cite{Keni2008,Milan2014,zhou2016large}: Recall, Multiple Object Tracking Accuracy (MOTA), Identity Switches (IDSW), Mostly Tracked (MT) and Mostly Lost (ML). MOTA considering false positives, false negatives and IDSW is a comprehensive tracking metric. In computing these metrics, we use a threshold of 20 pixels between the tracked fiber and the ground-truth fiber on each slice to count the hit/miss on the image. MT is the number of ground-truth fibers that are hit in no less than 80\% of slices while ML is the number of ground-truth fibers that are hit in no more than 20\% of slices. Higher (Recall, MOTA and MT) and lower (IDSW and ML) indicate the better tracking performance. 

\subsection{Results on fiber detection}
After the convergence of running Algorithm~\ref{alg1} on Set1 and Set2 respectively  (with tracking), we evaluate the large-scale fiber detection performance on the Set1 and Set2. Meantime, a well trained Faster R-CNN model $\mathbf{M}$ is obtained after the convergence of running Algorithm~\ref{alg1}. 

\begin{table*}[htbp]
	\begin{centering}
		\begin{tabular}{c||c|c|c|c|c|c}
			\hline
			\hline
			{Set1} & {with tracking?} & {Precision} & {Recall} & {F-measure} & {$N_{fp}$ per image} & {$N_{fn}$ per image} \tabularnewline
			\hline
			EdgeBox~\cite{zitnick2014edge} & no  & 93.0\% & 54.3\% & 68.6\% & 26.8 & 303.3 \tabularnewline
			ELSD~\cite{puatruaucean2012parameterless} & no    & 93.4\% & 92.5\% & 93.0\% & 43.1 & 49.3  \tabularnewline
			EMMPMH~\cite{zhou2016large}  &  no    & 96.9\% & 91.7\% & 94.2\% & 19.3 & 54.7 \tabularnewline
			Proposed-$\mathbf{M^{2}_{EdgeBox}}$ &  no   & 99.0\% & 97.3\% & 98.2\% & 6.1 & 17.5 \tabularnewline
			Proposed-$\mathbf{M^{2}_{EMMPMH}}$ &  no   & \textbf{99.3\%} & 96.1\% & 97.7\% & \textbf{3.9} & 28.3 \tabularnewline
			Proposed-EdgeBox &  yes   & 99.0\% & 93.2\% & 96.0\% & 6.0 & 45.1 \tabularnewline
			Proposed-EMMPMH &  yes   & 99.1\% & \textbf{98.1\%} & \textbf{98.6\%} & 5.2 & \textbf{12.2} \tabularnewline
			\hline
			\hline
			{Set2} & {with tracking?} & {Precision} & {Recall} & {F-measure} & {$N_{fp}$ per image} & {$N_{fn}$ per image} \tabularnewline
			\hline
			EdgeBox~\cite{zitnick2014edge} & no  & 92.0\% & 64.8\% & 76.0\% & 32.2 & 203.6 \tabularnewline
			ELSD~\cite{puatruaucean2012parameterless}  & no    & 91.0\% & 90.6\% & 90.8\% & 51.3 & 53.8  \tabularnewline
			EMMPMH~\cite{zhou2016large} & no     & 97.7\% & 95.1\% & 96.4\% & 12.5 & 28.2 \tabularnewline
			Proposed-$\mathbf{M^{1}_{EdgeBox}}$ & no & 98.8\% & 93.8\% & 96.2\% & 6.4 & 35.4 \tabularnewline
			Proposed-$\mathbf{M^{1}_{EMMPMH}}$  & no & 99.4\% & \textbf{98.5\%} & \textbf{99.0\%} & 3.1 & \textbf{8.6} \tabularnewline
			Proposed-EdgeBox & yes & 99.3\% & 98.4\% & 98.9\% & 3.9 & 8.7 \tabularnewline
			Proposed-EMMPMH  & yes & \textbf{99.5\%} & 98.1\% & 98.8\% & \textbf{2.4} & 10.6 \tabularnewline
			\hline
			\hline
			{Set3} & {with tracking?} & {Precision} & {Recall} & {F-measure} & {$N_{fp}$ per image} & {$N_{fn}$ per image} \tabularnewline
			\hline
			EdgeBox~\cite{zitnick2014edge}  & no  & 93.4\% & 56.0\% & 70.1\% & 24.2 & 270.6 \tabularnewline
			ELSD~\cite{puatruaucean2012parameterless} & no  & 93.6\% & 91.8\% & 92.7\% & 38.4 & 50.5  \tabularnewline
			EMMPMH~\cite{zhou2016large}  & no  & 97.2\% & 97.7\% & 97.4\% & 17.2 & 14.1 \tabularnewline
			Proposed-$\mathbf{M^{1}_{EdgeBox}}$ & no & 99.1\% & 93.5\% & 96.2\% & 4.7 & 39.9 \tabularnewline
			Proposed-$\mathbf{M^{1}_{EMMPMH}}$  & no & 99.3\% & 98.5\% & 98.9\% & 3.8 & 9.0 \tabularnewline
			Proposed-$\mathbf{M^{2}_{EdgeBox}}$ & no & 99.0\% & \textbf{99.0\%} & \textbf{99.0\%} & 5.8 & \textbf{6.0} \tabularnewline
			Proposed-$\mathbf{M^{2}_{EMMPMH}}$  & no & \textbf{99.5\%} & 98.5\% & \textbf{99.0\%} & \textbf{2.5} & 8.7 \tabularnewline
			\hline
			\hline
		\end{tabular}
		\par\end{centering}
	\centering{}\caption{Large-scale fiber detection performance. Set1 and Set2 are two image sequences and Set3 is a set of single images. Note that each image contains about 600 fibers and all the methods in this table are unsupervised.}
	\label{tab:detection}
\end{table*}

\begin{figure*}[htbp]
\begin{centering}
\includegraphics[width=1.0\textwidth]{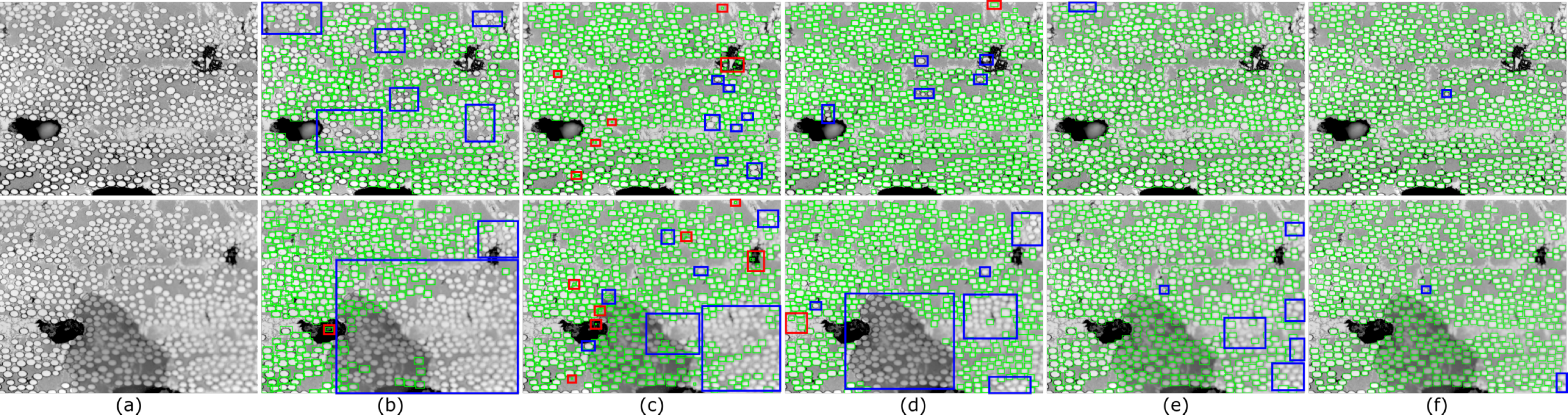}
\par\end{centering}
\caption{Illustration of large-scale fiber detection. (a) two sample images (clear and degraded), and detections by (b) EdgeBox, (c) ELSD, (d) EMMPMH, (e) Proposed-$\mathbf{M^{1}_{EdgeBox}}$ and (f) Proposed-$\mathbf{M^{1}_{EMMPMH}}$. Fibers are detected as green bounding boxes. Red and blue boxes highlight the false positive and false negative errors respectively.}
\label{fig:detection-demo}
\end{figure*}

Using the well trained Faster R-CNN model $\mathbf{M}$, we directly apply it to detect the large-scale fibers on another two datasets (without tracking). The performance on Set1, Set2 and Set3 is summarized in Table~\ref{tab:detection}. We can see that Proposed method using EMMPMH as initialization achieves the best performance in most cases with high Precision, Recall and F-measure and low $N_{fp}$ per image and $N_{fn}$ per image. The Proposed method using EdgeBox as initialization achieves second best performance and comparable or better performance than the state-of-the-art algorithm EMMPMH. Even without using any manual annotations, the proposed method could achieve nearly 99\% F-measure for large-scale fiber detection on three datasets, which fully demonstrates the accuracy and effectiveness of the proposed method. The fiber detection example is shown in Fig.~\ref{fig:detection-demo}. 

\subsection{Improvement from initialization and algorithm convergence}
\begin{figure*}[htbp]
\begin{centering}
\includegraphics[width=1.0\textwidth]{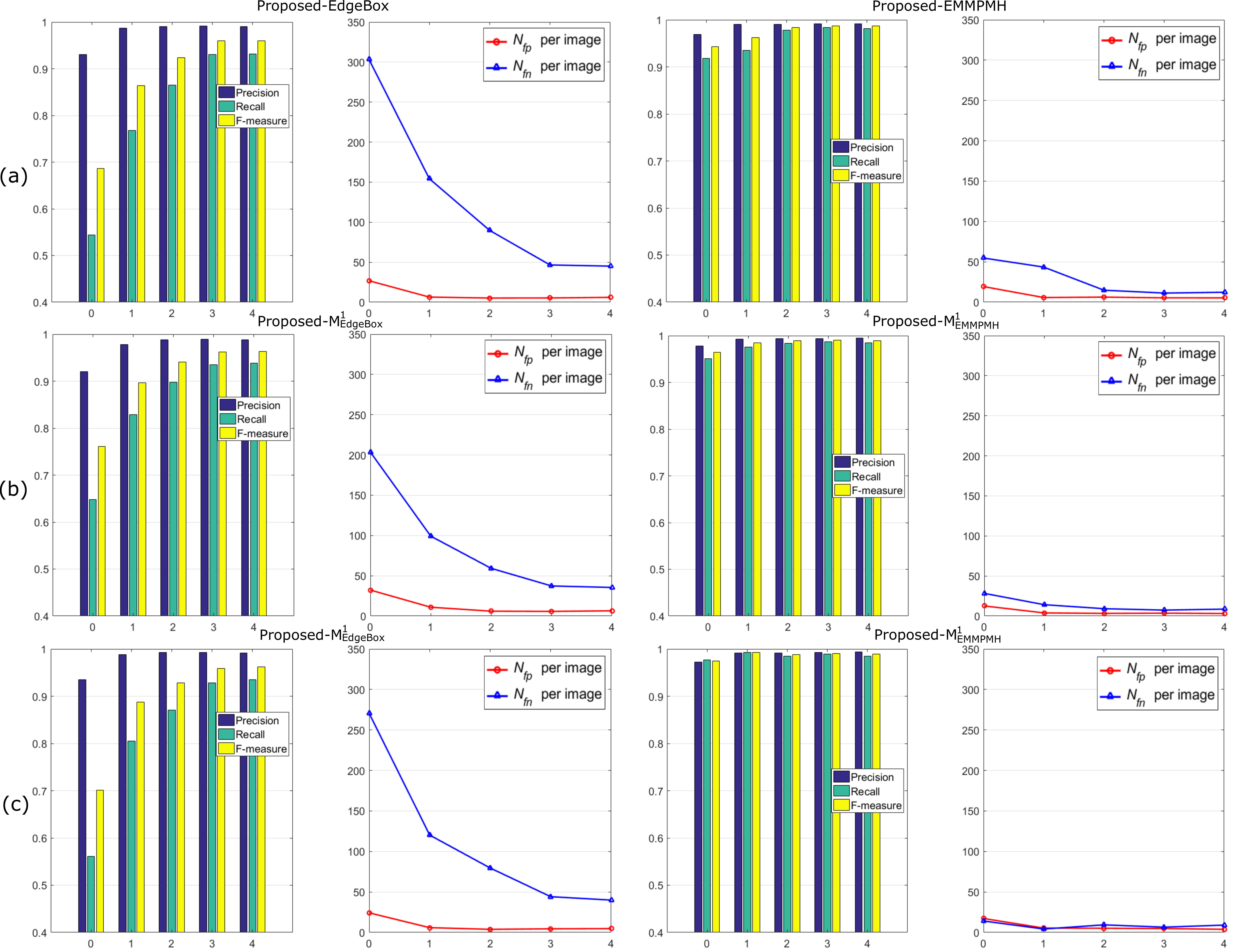}
\par\end{centering}
\caption{Illustration of the fiber-detection performance change of the proposed method in (a) Set1, (b) Set2, and (c) Set3 from iteration 0 to iteration 4. Left two columns display the proposed method using EdgeBox as initialization and right two columns show the proposed method using EMMPMH as initialization.}
\label{fig:detection-convergence}
\end{figure*}

In this section, we will show the improvement from initialization and the algorithm convergence. We treat initialization result as the iteration 0, and then show the fiber-detection performance change from iteration 0 to iteration 4 in Fig.~\ref{fig:detection-convergence}. We can see that the proposed method significantly improve the fiber-detection performance by both the initialization EMMPMH and EdgeBox. From iteration 0 to 4 using either initialization, the proposed method could incrementally boost the Precision, Recall and F-measure, and simultaneously reduce the $N_{fp}$ per image and $N_{fn}$ per image. We can see that the proposed method converges in 3 to 4 iterations.    

\subsection{Results on fiber tracking}
For tracking-by-detection algorithms, previous study~\cite{hong2016online} has shown that the performance of object tracking is highly dependent on the performance of object detection, especially the recall performance. In this section, we will show the performance change of large-scale fiber tracking in different iterations by Proposed-EdgeBox and Proposed-EMMPMH on Set1. Because it is hard to manually annotate the ground-truth trajectories of all the fibers, we follow the same strategy in~\cite{yu2016groupwise, zhou2016large} to prune unrelated trackers for evaluation. The tracking performance change is summarized in Table~\ref{tab:tracking}. From iteration 0 to 4 using either initialization, the proposed method could incrementally boost the tracking performance. The proposed method could obtain better tracking performance if better initialization is applied. Compared to the baseline performance by tracking initialized detections (iteration=0), the tracking performance is greatly improved, with either EdgeBox or EMMPMH as initializations.             
 
\begin{table}[htbp]
	\footnotesize
	\begin{centering}
		\begin{tabular}{c||c|c|c|c|c}
			\hline
			{Proposed-EdgeBox} & {Recall} & {MOTA} & {IDSW} & {MT} & {ML} \tabularnewline
			\hline
			iteration=0   & 46.6\% & 41.0\% & 1666 & 129 & 168 \tabularnewline
			iteration=1   & 89.6\% & 88.3\% & 123 & 412 & 29  \tabularnewline
			iteration=2   & 92.6\% & 91.9\% & 56 & 436 & 23 \tabularnewline
			iteration=3   & 94.4\% & 94.0\% & 58 & 451 & 21 \tabularnewline
			iteration=4  & \textbf{95.2\%} & \textbf{94.6\%} & \textbf{47} & \textbf{452} & \textbf{13} \tabularnewline
			\hline
			\hline
			{Proposed-EMMPMH} & {Recall} & {MOTA} & {IDSW} & {MT} & {ML} \tabularnewline
			\hline
			iteration=0   & 92.2\% & 91.0\% & 96 & 435 & 24 \tabularnewline
			iteration=1   & 96.5\% & 95.8\% & 81 & 465 & 12  \tabularnewline
			iteration=2   & 97.2\% & 96.7\% & 52 & 467 & 9 \tabularnewline
			iteration=3   & 98.3\% & 98.1\% & 17 & 471 & 6 \tabularnewline
			iteration=4  & \textbf{99.3\%} & \textbf{99.3\%} & \textbf{4} & \textbf{477} & \textbf{1} \tabularnewline
			\hline
		\end{tabular}
		\par\end{centering}
	\centering{}\caption{Large-scale fiber tracking performance on Set1. We manually annotate 481 fibers' trajectories as the tracking ground truth for evaluation.}
	\label{tab:tracking}
\end{table}  
\section{Conclusions}\label{sec:conclusions}
In this paper, we proposed an unsupervised method to detect and track large-scale fibers in microscopic material images. The proposed method alternately run Faster R-CNN and fiber tracking algorithm in several iterations to improve fiber detection and tracking simultaneously. The proposed method takes an image sequence as input without requiring any manual annotations, achieving nearly 99\% F-measure for fiber detection and 99\% MOTA for fiber tracking. A well trained Faster R-CNN model is obtained by the proposed method, and then we apply it to detect large-scale fibers in single images of   different datasets and also achieved nearly 99\% F-measure as detection performance. The experimental results show that the proposed unsupervised method is accurate and effective in detecting large-scale fibers, leading to improved fiber tracking. 

{\small
\bibliographystyle{ieee}
\bibliography{egbib}
}

\end{document}